\documentclass{article}

\usepackage{PRIMEarxiv}

\usepackage[utf8]{inputenc} 
\usepackage[T1]{fontenc}    
\usepackage{hyperref}       
\usepackage{url}            
\usepackage{booktabs}       
\usepackage{amsfonts}       
\usepackage{nicefrac}       
\usepackage{microtype}      
\usepackage{lipsum}
\usepackage{fancyhdr}       
\usepackage{graphicx}       
\graphicspath{{media/}}     
\usepackage{threeparttable}

\title{Deciphering Diagnoses: How Large Language Models Explanations Influence Clinical Decision Making
}

\author{
  D. Umerenkov \\
  Sber AI Lab \\
  \texttt{d.umerenkov@gmail.com} \\
   \And
  G. Zubkova \\
  Sber AI Lab \\
  \texttt{galina.v.zubkova@gmail.com } \\
  \And
   A. Nesterov \\
  Sber AI Lab \\
  \texttt{ainestetov@sberbank.ru} \\
}

\begin{document}
\maketitle

\begin{abstract}
Clinical Decision Support Systems (CDSS) utilize evidence-based knowledge and patient data to offer real-time recommendations, with Large Language Models (LLMs) emerging as a promising tool to generate plain-text explanations for medical decisions. This study explores the effectiveness and reliability of LLMs in generating explanations for diagnoses based on patient complaints. Three experienced doctors evaluated LLM-generated explanations of the connection between patient complaints and doctor and model-assigned diagnoses across several stages. Experimental results demonstrated that LLM explanations significantly increased doctors' agreement rates with given diagnoses and highlighted potential errors in LLM outputs, ranging from 5\% to 30\%. The study underscores the potential and challenges of LLMs in healthcare and emphasizes the need for careful integration and evaluation to ensure patient safety and optimal clinical utility.
\end{abstract}


\section{Introduction}

Clinical Decision Support Systems (CDSS) are tools aiding healthcare professionals by using evidence-based knowledge and patient data to provide real-time recommendations. Integrated with Electronic Health Records (EHR), they assist in diagnosis, treatment recommendations, and predicting patient outcomes. As healthcare evolves, the significance of CDSS in patient-centric care grows, highlighting its necessity in current clinical processes. 
Explainability in CDSS is vital, yet its absence may lead to reliance issues. Clear explanations behind AI recommendations are essential for informed clinical decisions. The need for explainable AI (XAI) is emphasized by the call for ethical decision-making and understanding that AI can reflect historical biases, stressing transparency's role.

Assessing explainability in medical AI is complex. The term 'explainability' varies per stakeholder, and achieving the right level of detail is challenging. In healthcare's critical context, the lack of standardized evaluation methods for explainability remains a significant concern.
In this work, we explore explaining the connection between patients' complaints entered into the EHR system and the diagnoses provided by the attending physician or by an AI model trained exclusively on patients' complaints.

While diverse modalities of explanations are available, the recent advancements in Large Language Models (LLMs) enable us to explore plain-text explanations. Such interpretations are most intuitive for users if we can maintain a high enough quality standard.
LLMs have shown their versatility by addressing a plethora of medical tasks, including but not limited to image captioning \cite{selivanov2023medical}, text summarization \cite {madden2023assessing}, and answering complex queries \cite{mathur2023summqa}. In the realm of healthcare, LLMs have the potential to transform the way we interpret and communicate complex medical data. While LLMs can generate explanations, they are not infallible. Like any model, they may produce inaccuracies or miss nuances essential in a medical context. These potential errors or inaccuracies in explanations provided by LLMs necessitate thorough evaluation. Human expertise becomes indispensable in discerning and quantifying these errors, especially in such critical domains as healthcare.

We organized a series of experiments designed to gauge the quality and precision of the explanations produced by LLMs, evaluate the impact of these explanations on medical professionals' decision-making, and discern disparities between explanations for diagnoses made by humans and those by the model. Given the tempo of technological advancements, we foresee LLMs soon playing a pivotal role in offering medical explanations to a broad audience, from novices to seasoned medical experts. This research aims to start bridging the current gaps and ensure the smooth incorporation of LLMs into the CDSS framework.

\section{Related work}

Clinical Decision Support Systems (CDSS) have garnered significant attention over the last few decades as essential tools in enhancing patient care and improving healthcare delivery efficiency. CDSS provide healthcare professionals with knowledge and patient-specific information to enhance decision-making at the point of care \cite{osheroff2007roadmap}. These systems incorporate evidence-based knowledge repositories and patient data to offer tailored recommendations, which can significantly reduce medical errors, improve healthcare outcomes, and reduce costs \cite{roshanov2011computerized}. While the adoption of CDSS promises substantial advantages, challenges such as the need for seamless integration with Electronic Health Records (EHR), potential alert fatigue, and concerns about data privacy and accuracy have been highlighted \cite{bates2003ten}. However, as artificial intelligence and machine learning techniques continue to evolve, there is potential for even greater personalization and accuracy in the recommendations provided by CDSS \cite{shortliffe2018clinical}.

In recent years, the incorporation of Explainable Artificial Intelligence (XAI) into Clinical Decision Support Systems (CDSS) has gained significant attention, primarily due to the critical nature of medical decisions and the ethical necessity of interpretability. Amann et al. in \cite{amann2020explainability} highlight the multifaceted nature of the challenges involved in AI explainability in healthcare, emphasizing the crucial balance between model performance and interpretability, and the need for collaboration across domains such as medicine, computer science, and ethics. This underscores the argument that mere algorithmic transparency may not suffice in clinical contexts. Similarly, Antoniadi et al. in \cite{antoniadi2021current} elaborate on the existing hurdles XAI faces in CDSS, such as inconsistency in explanations, potential over-reliance on AI insights, and the possible cognitive burden on healthcare professionals. This review also underscores the potential of XAI in improving trustworthiness, user adoption, and, ultimately, patient outcomes in CDSS. Together, these studies reflect an emerging consensus on the imperativeness of XAI in the medical domain, urging researchers and practitioners to navigate the nuanced challenges of integrating explainability into healthcare AI solutions.

\cite{rajkomar2019machine} emphasized the essentiality of transparency in healthcare AI, highlighting the dangers of black-box models that may offer predictions without any tangible reasoning. These challenges are partially addressed by techniques such as LIME (Local Interpretable Model-agnostic Explanations) \cite{ribeiro2016should}, a method that explains predictions of any classifier by approximating it with an interpretable model locally. In \cite{lundberg2017unified} introduced SHAP (SHapley Additive exPlanations), an approach to explain any machine learning model's output by combining several methods based on cooperative game theory. This methodology has been used in medical applications for better interpretability. In \cite{vaswani2017attention}, authors introduced the self-attention mechanism that is the basis of the LLMs, and it can be used to infer what parts of the input text had the most influence on the model predictions. Though these methods provide valuable information about how the model arrived at the conclusions, their output is not trivial to interpret, which provides unique challenges. The best way to explain the model results would be a plain text explanation understandable by humans, which is the focus of this work.

In \cite{panigutti2022understanding} authors experimented on how the provision of explanation influences the medical expert attitude towards the AI model predictions of the chances of a patient shortly suffering from an acute myocardial infarction. 
In \cite{du2022role} authors explore how the provision of explanation by feature contribution (SHAP features) and explanation by example affects the prediction of gestational diabetes mellitus by a SVM model by healthcare practitioners.
In \cite{lee2023understanding} authors assess the effect of providing salient and counterfactual features based on SHAP features on doctors' agreement with model predictions on assessing the quality of motion of patients affected by stroke.
In all these works, the explanation provision statistically significantly affected doctor agreement rates but needed a custom-tailored interface to present the explanations.

There is much recent work regarding integrating LLMs into CDS systems; see, for example \cite{lee2023benefits}. For a general overview of LLMs applications in medicine, see \cite {muftic2023exploring}.
In \cite{li2023explaincpe} authors used ChatGPT and GPT-4 to generate and explain the multiple-choice questions from the Chinese farmacaeutical exam.
In this work, we follow the general direction outlined in \cite{hassija2023interpreting} as "Once trained, the GPT model can generate explanations in natural language for specific black-box models based on their inputs and outputs." by using the GPT model to explain the correspondence between patients symptoms and its diagnosis.

\section{Experimental setup}

In this work, we used the data from the train split of the RuMedBench dataset \cite{RuMedBench}. Each entry consists of patient complaints entered by the doctor into the electronic health record and the ICD code of the diagnosis assigned by the doctor. This ICD code is referred to hereafter as ground truth data. This dataset is in Russian, but we present translated examples in the article. We included the original examples in Russian in the supplementary materials.

The explanations for symptoms-diagnosis correspondence were generated by \emph{gpt-3.5-turbo model} via API at \emph{https://api.openai.com/v1/chat/completions} using the following prompt in Russian \emph{"I have complaints SYMPTOMS and the diagnostician assigned the diagnosis DIAGNOSISNAME. Explain why my complaints correspond to DIAGNOSISNAME."}

In these experiments, three Russian-speaking doctors participated: a Therapist/Cardiologist with 40 years of experience, hereafter referred to as Doctor 1; a Therapist/Cardiologist with 9 years of experience, hereafter referred to as Doctor 2; and a Therapist with 28 years of experience, hereafter referred to as Doctor 3. We did not provide the doctors with any information regarding the source of the data or the source of provided explanations.

We have conducted the experiments in three stages. All three doctors completed each stage before receiving the data for the next stage.

\subsection{First stage - evaluating the explanation quality for ground truth data}

The doctors were provided with 500 pairs of complaints-diagnoses explanations and were asked to answer three questions
\begin{itemize}
\item Is the presented diagnosis a valid diagnostic hypothesis? 
\item Is the presented text an explanation of why the patient complaints lead to the presented diagnosis?
\item Does the explanation contain errors of any kind?
\end{itemize}
The doctors answered the second and third questions only for cases with affirmative answers to the first question. If the doctor disagreed with the provided diagnosis, he did not evaluate the correctness of the explanation.

\subsection{Second stage - evaluating the effect of an explanation on doctor agreement with the provided diagnosis}

We provided the doctors with 500 pairs of complaints-diagnoses without providing an explanation. Of these 500 pairs, 473 were the same as at the first stage. We again asked the doctors if the presented diagnosis was a valid diagnostic hypothesis as the only question at this experiment stage.

\subsection{Third stage - evaluating the explanation quality for model predictions}
At this stage, we used a transformer classification model described in detail \cite{blinov2020predicting} to generate new diagnoses for each of the 500 patients' complaints from the first stage. These diagnoses are referred to hereafter as model predictions. After generating new diagnoses, we generated new explanations as described above. The examples were randomly shuffled at this stage. We asked the doctors the same three questions as in the first stage.

\section{Experiment results}

\subsection{Effect of  the explanation on doctor agreement with the provided diagnosis}

To evaluate the effect of automatically generated explanations on doctors' agreement with the diagnostic hypothesis, we compared the rates of doctors' agreement with the provided diagnosis in the presence of an explanation during stage 1 and in the absence of such an explanation during stage 2. The results are presented in Table \ref{tab:table_agreement}.
There were cases for all the doctors when the provision of an explanation led to disagreement with the diagnosis. However, such cases were considerably rarer than when such provisions led to agreement.
Overall, our data show that the provided explanation led to an increased agreement rate with the provided diagnosis for all three doctors, and this increase was statistically significant. The statistical significance was calculated using an exact p-value binomial test.

\begin{table}
 \caption{Effect of LLM explanation on doctor agreement with diagnostic hypothesis }
  \centering
  \begin{tabular}{llll}
    \toprule
         &  Doctor 1 & Doctor 2  & Doctor 3  \\
    \midrule
    Agree with explanation & 426 & 397 & 455   \\
    Agree without explanation & 399 & 378 & 414      \\
    Agree in any case & 388 & 357 & 407  \\
    Disagree in any case & 42 & 60 & 17  \\
    Agree only with explanation & 38 & 39 & 48  \\
    Agree only without explanation & 11 & 17 & 7  \\
    \midrule
    p-value & 0.0001 & 0.0046 & 0.0000  \\
    \bottomrule
  \end{tabular}
  \label{tab:table_agreement}
\end{table}

\subsection{Doctor agreement with the ground truth diagnosis vs. model diagnosis}

We also checked what diagnoses were more agreeable to the doctors, those provided by the model or those from the dataset. The number of cases where the model diagnosis coincided with ground truth diagnosis was relatively small - only 131 cases out of 500.
The results are presented in Table \ref{tab:table_gt_vs_model}. Surprisingly, the doctors agreed more with the diagnoses provided by the model than the ones entered by the doctor. Furthermore, this difference was statistically significant for two doctors out of three. We hypothesize that this is not an effect of the model being a better diagnostician than the doctors but rather because the GT diagnosis was made based on more information than is contained in the patient's complaints. The absence of this information leads to diagnoses that are not directly tied to the provided complaints. On the other hand, the model diagnoses were made based only on the patient complaints and, therefore, aligned better with the presented data.

\begin{table}
 \caption{GT diagnoses vs predicted diagnoses }
  \centering
  \begin{tabular}{llll}
    \toprule
         &  Doctor 1 & Doctor 2  & Doctor 3  \\
    \midrule
    Total cases with valid answers & 499 & 491 & 500   \\
    Agree with GT & 444 & 411 & 475      \\
    Agree with MODEL & 470 & 438 & 484      \\
    Agree in any case & 422 & 375 & 460  \\
    Disagree in any case & 7 & 17 & 1  \\
    Agree only with GT & 22 & 36 & 15  \\
    Agree only with MODEL & 48 & 63 & 24  \\
    \midrule
    p-value & 0.0025 & 0.0086 & 0.1996
  \\
    \bottomrule
  \end{tabular}
  \label{tab:table_gt_vs_model}
\end{table}

\subsection{Explanation quality for ground truth and model predictions}

To evaluate the quality of the provided explanation, we first filtered our results to include only cases where all three doctors agree with the presented diagnosis. After such filtering, we were left with 388 cases for ground truth predictions and 425 cases for model predictions. The results are presented in Table \ref{tab:table_explanation_acceptance} and show that the acceptance ratio is similar across both settings and varies from 76\% to 98\%

\begin{figure}
     \centering     
     \includegraphics[width=400pt]{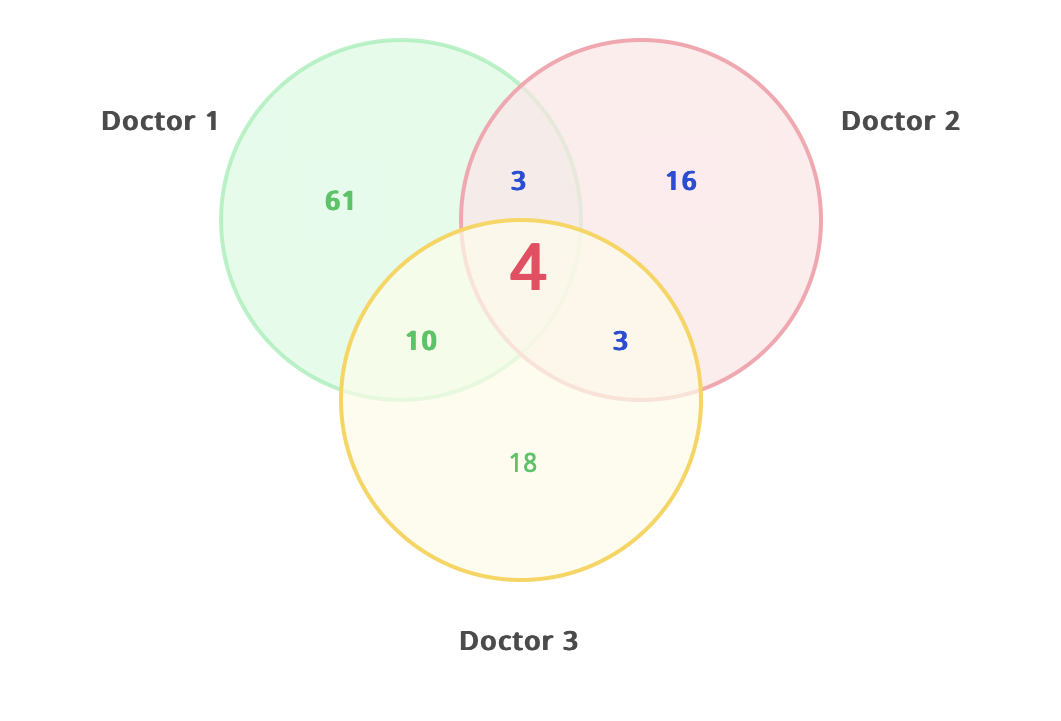}
     \caption{Inter-doctor agreement on what explanations are erroneous for GT predictions}
     \label{fig:GT_interagree}
\end{figure}

We also checked for overlap in the cases where doctors flagged the explanation as incorrect, and such overlap was surprisingly small. The results are presented in Figure \ref{fig:GT_interagree} for the ground truth case and Figure \ref{fig:GT_interagree} for the model prediction case. Considering incorrect the cases where the majority of the doctors (two or three) flagged the case as incorrect, we arrive at an error rate of 5\% (20 of 388) cases for ground truth setting and 4\% (20 of 445) cases for model prediction setting. The detailed analysis of the cases showed that even the errors marked by only one of the doctors often raised valid concerns missed by the other two doctors. If we consider incorrect the cases where at least one of the doctors raised concerns, the number of erroneous case rise to 30\% (115 of 388) in the ground truth setting and to 27\% (122 of 445) in the model prediction setting. We hypothesize that the actual number of incorrect cases is somewhere in 5\%-30\%.

\begin{table}
 \caption{LLM explanation acceptance by doctors}
  \centering
  \begin{tabular}{llll}
    \toprule
         &  Doctor 1 & Doctor 2  & Doctor 3  \\
    \midrule
    \multicolumn{4}{c}{Ground truth - 388 cases} \\
    \midrule
    Text is explanation & 343 (88\%) & 387 (100\%) & 388 (100\%)   \\
    Text does not contain errors & 311 (80\%) & 362 (93\%) & 353 (91\%)     \\
    Text is explanation and does not contain errors & 310 (80\%) & 362 (93\%) & 353 (91\%)  \\
    \midrule
    \multicolumn{4}{c}{Model predictions - 425 cases} \\
    \midrule
    Text is explanation & 388 (91\%) & 425 (100\%) & 423 (100\%) \\
    Text does not contain errors & 323 (76\%) & 388 (91\%) & 420 (99\%)    \\
    Text is explanation and does not contain errors & 322 (76\%) & 388 (91\%) & 418 (98\%)       \\
    \bottomrule
  \end{tabular}
  \label{tab:table_explanation_acceptance}
\end{table}

\begin{figure}
     \centering     
     \includegraphics[width=400pt]{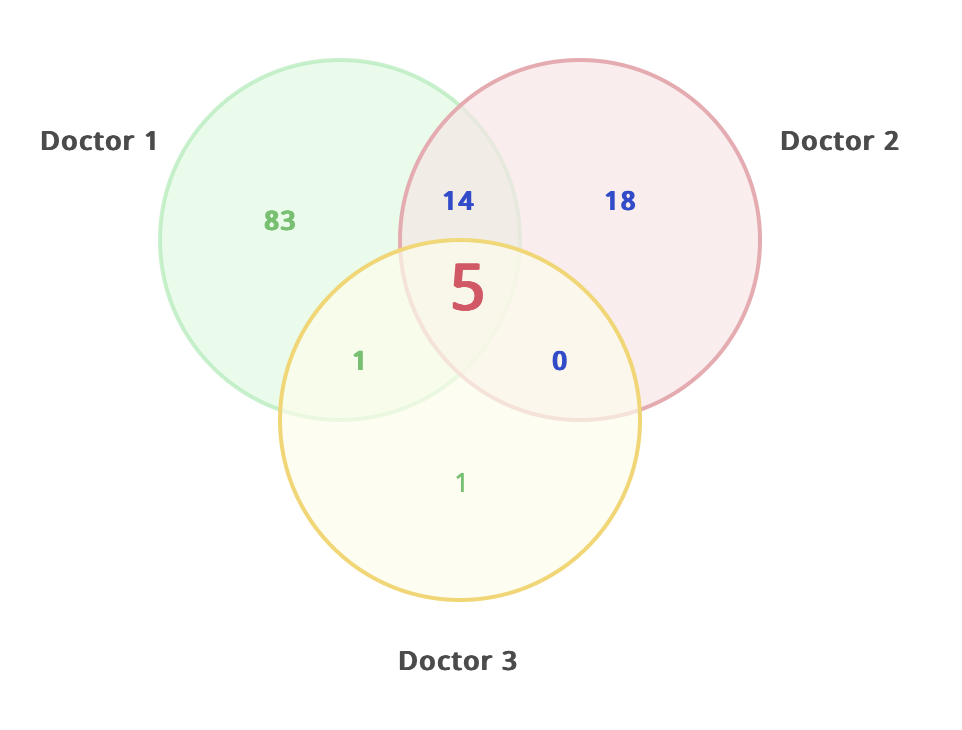}
     \caption{Inter-doctor agreement on what explanations are erroneous for MODEL predictions}
     \label{fig:MODEL_interagree}
\end{figure}

\subsection{Experiment consistency evaluation}

We used the 131 cases with the same diagnoses in stages 1 and 3 to check for the reproducibility of the results and measure the noise inherent in doctor evaluation. We presented these cases to the doctors in a different order and with a time delay of over a month, so the doctor made the decision anew in each case. Also, due to the stochastic nature of LLM predictions, the presented explanations were different, while the diagnoses were the same.
In the absence of any noise, we would expect that provided the same set of complaints and diagnosis, the doctors would give the same answer, but this is not the case, as seen in table \ref{tab:table_interagreement}. The number of cases where the doctor gave the same answer on diagnosis agreement was 93\%-98\%, and the correctness of the explanation was 72\%-92\%

\begin{table}
 \caption{Interagreement between experiment stages 1 and 3}
  \centering
  \begin{tabular}{llll}
    \toprule
         &  Doctor 1 & Doctor 2  & Doctor 3  \\
    \midrule
    \multicolumn{4}{c}{Same diagnosis - 131 cases} \\
    \midrule
    Is valid diagnostic hypothesis & 97\% & 93\% & 98\%   \\
    \midrule
    Text is explanation & 87\% & 93\% & 82\%   \\
    Text does not contain errors & 72\% & 82\% & 92\%     \\

    \bottomrule
  \end{tabular}
  \label{tab:table_interagreement}
\end{table}

\section {Error analysis}

In order to analyze the errors of a model, one of the authors with medical expertise and 7 years of experience analyzed 40 of the cases; each flagged as incorrect by at least two of the medical experts. During the analysis, the comments of the medical experts were classified according to the types of errors in the explanation.

The four types of errors were:
\begin{itemize}
\item Inventing connections between symptoms and conditions that do not exist
\item Inventing symptoms that were not present in the patients' complaints
\item Vague or unclear argumentation
\item Model refusal to make an explanations
\end{itemize}

The analysis results are presented in Table \ref{tab:table_error_layout}; Selected examples are presented in Table \ref{tab:table_error_examples} and full analysis data is available in supplementary materials.

\begin{table}
 \caption{Types of errors made by the model}
  \centering
  \begin{tabular}{lll}
    \toprule
         &  Doctors evaluation & Our evaluation  \\
    \midrule
    Inventing connections between symptoms and conditions & 30 & 23   \\
    Inventing symptoms & 12 & 24     \\
    Vague or unclear argumentation & 11 & 21    \\
    Disagreement with the diagnostic hypothesis & 6 & 11  \\
    Model refusal to make an explanations & 1 & 1  \\
    \bottomrule
  \end{tabular}
  \label{tab:table_error_layout}
\end{table}

The errors identified by the authors differed significantly from those identified by the medical experts. Additionally, the authors identified substantially more errors. We believe that to be a result of both a more thorough analysis of fewer cases by the author and familiarity with the kind of mistakes common for LLMs.

\begin{table}
\begin{threeparttable}[b]
 \caption{Examples of errors made by the model}
  \centering
  \begin{tabular}{p{2cm}p{1.5cm}p{8.2cm}p{3.2cm}}
    \toprule
     Complaints & Diagnosis & Explanation & Doctors comments \\
    \midrule
     Pain in the left ear area, sensation of heaviness and lump in the throat (only occurs in the morning hours), skin pain (left half of the face, hand, tingling in this area). Painful to swallow. Anxiety. & Disorders of autonomic nervous system & The diagnosis of autonomic nervous system disorder can be explained by your complaints for the following reasons: \newline Pain in the ear area and lump in the throat: The autonomic nervous system regulates sympathetic and parasympathetic nerve fibers that control the functions of the head and neck organs, including the ear and throat. Dysfunction of this system can lead to altered pain perception and the emergence of vague symptoms like pain, sensation of heaviness, and lumpiness. \newline Skin pain and tingling in the left half of the face and hand: An autonomic nervous system disorder can cause changes in the function of sympathetic nerve fibers that control blood supply and skin innervation. This can lead to the emergence of painful sensations, skin color change, and tingling. 3. Pain when swallowing: The autonomic nervous system also controls swallowing function through the influence of parasympathetic innervation of the pharynx and pharyngeal muscles. Dysfunction of this system can lead to coordination issues of pharyngeal movements and cause pain during swallowing. \newline Anxiety: The autonomic nervous system plays a crucial role in regulating emotional state. Dysfunction of this system can cause an imbalance in the neurochemical processes associated with anxiety, leading to anxious symptoms. \newline However, it should be noted that an accurate diagnosis requires additional tests and consultation with a specialist. It is recommended to consult a doctor for a detailed discussion of your symptoms and their causes. & 1. Pain syndrome in the ear and when swallowing are not typical symptoms of an autonomic system disorder. \newline 3. The diagnosis cannot be explained by such complaints. \\
    \midrule
     Pains in the sternum area, moderately expressed, numbness of the tongue, face, numbness of the hands, lower extremities, elevated blood pressure at 160/70 mmHg, shortness of breath when walking. & Hypertensive heart disease & Your complaints correspond to the diagnosis of hypertensive heart disease with predominant heart damage for the following reasons: \newline Chest pain can be caused by narrowing or spasm of the coronary arteries, which is a characteristic manifestation of hypertensive heart disease. High blood pressure can place an increased strain on the heart, leading to ischemia and chest pains. \newline Numbness of the tongue, face, hands, and lower extremities can be related to circulatory disturbances due to elevated blood pressure. This might result from damage to vessels and nerve endings in the areas of the head, face, and extremities. \newline Elevated blood pressure (160/70 mmHg) is a characteristic feature of hypertensive heart disease. High pressure on arterial walls can lead to their damage and thickening, which in turn exacerbates hypertension and promotes the development of cardiovascular complications. \newline Shortness of breath when walking is a result of deteriorating heart function caused by hypertension. High blood pressure can lead to left ventricular dysfunction, compromising its ability to effectively contract and pump blood. This can cause shortness of breath during physical activity. \newline However, I am not a qualified diagnostician, and I recommend that you see a doctor for a comprehensive medical examination and to receive detailed consultation and treatment. & 1. Pains behind the sternum are more characteristic of ischemic heart disease. Numbness of the tongue and extremities requires the exclusion of neurological pathology. \newline 3. Numbness of the tongue, face, and hands are symptoms of focal neurological symptomatology with hypertension. It would have been excellent if the phrase included a disruption in cerebral circulation; that would have been ideal.. \\

    \bottomrule
  \end{tabular}
      %
  \label{tab:table_error_examples}
  \end{threeparttable}[b]
\end{table}

\begin{table}
 \caption{Examples of errors made by the model (continued)}
  \centering
  \begin{tabular}{p{2cm}p{1.5cm}p{8.2cm}p{3.2cm}}
    \toprule
     Complaints & Diagnosis & Explanation & Doctors comments \\
    \midrule
     Complaints of bruising on the body when measuring blood pressure from the cuff for 3 days. Blood pressure (BP) control shows 113-140/66-82 and heart rate (HR) 60-72. & Hypertensive heart disease & The diagnosis 'Hypertensive heart disease' matches your complaints, as bruises on the body can be associated with circulatory disorders and increased pressure in the vessels. Hypertensive heart disease is characterized by high blood pressure, which can cause vascular disorders and damage to organs, including the heart. In this case, predominant heart damage may indicate that the cardiac muscle suffers from prolonged high pressure, which can lead to the appearance of bruises. With such a diagnosis, it is important to monitor blood pressure and heart rate to prevent further damage to the heart and other organs. & 1. The appearance of subcutaneous hematomas on the skin is not related to high blood pressure. \newline 3. That may be the case, but the complaints show normal pressure, and even with high BP, subcutaneous hemorrhages do not occur. This means that these are manifestations of another disease (condition), for example, some form of thrombocytopenia or hemorrhages. \\
    \midrule
     General weakness, decreased work capacity, episodic pain in the right hypochondrium. & Other forms of hypothyroidism & Other forms of hypothyroidism include any forms of hypothyroidism other than primary and secondary hypothyroidism. Typically, they are associated with a reduction in the amount of thyroid hormones, which is not due to a malfunction of the gland itself. This can be caused by various factors, such as immune system disorders, Sjögren's syndrome, cancer, and other diseases, as well as radiation therapy treatment and medications like amiodarone and lithium. These factors can lead to general weakness, decreased work capacity, and other symptoms described in the patient's complaints. Therefore, the diagnosis "Other forms of hypothyroidism" matches his complaints. & 3. Weakness is a non-specific symptom that can be present in any disease and in hypothyroidism, except for the pain in the right hypochondrium. \\ 
    \midrule
     Belching, burning in the epigastrium, sleep disturbances, intolerance to bakery products. & Gastritis and duodenitis & Complaints of belching, burning in the epigastrium, sleep disturbances, and intolerance to bakery products are the main symptoms of gastritis and duodenitis - inflammatory diseases of the mucous membranes of the stomach and duodenum respectively. Gastritis and duodenitis can be caused by various factors: improper diet, stress, Helicobacter pylori infection, among others. With gastritis and duodenitis, there's a disruption in the functioning of the gastrointestinal tract, which manifests as belching, burning in the epigastrium, sleep disturbances, and intolerance to certain foods. Therefore, the symptoms described in the complaints correspond to the diagnosis of gastritis and duodenitis. & 2. May indicate lactose intolerance or insufficiency of the cardia (or gastroesophageal junction). \newline 3.Intolerance to bakery products is NOT the main symptoms of gastritis and duodenitis; they are the primary symptoms of gluten intolerance (celiac disease), where there may be stomach symptoms, but primarily symptoms of intestinal dyspepsia. \\ 
    \bottomrule
  \end{tabular}
  \label{tab:table_error_examples}
\end{table}

Overall, the majority of cases analyzed had several groups of complaints, some of which were related to the diagnosis and others were not. The model, however, attempted to link all symptoms to the diagnosed condition, thereby causing most errors. Errors related to "inventing" non-existent facts likely stemmed from the model's overzealous efforts to explain the diagnosis. The medical experts focused on checking if certain symptoms corresponded to specific diagnoses; however, they overlooked the factual inaccuracies made by the model. Some cases were difficult to diagnose based on the presented symptoms alone, but the model still attempted to explain the connection between symptoms and diagnosis.

\section{Discussion}

Clinical Decision Support Systems (CDSS) using advanced machine learning models, particularly Large Language Models (LLM), have the potential to revolutionize the delivery and quality of healthcare. This research reveals that while these models can generate influential explanations, there are caveats that stakeholders must address to ensure patient safety and optimize clinical utility.

\subsection{Evaluating Explanation Quality}

One of the significant challenges highlighted in this research is the inherent complexity in defining and evaluating the quality of explanations. While our research deployed LLM to generate explanations for medical diagnoses, the term 'quality' proved elusive, as evidenced by low inter-agreement among our participating medical professionals. The differences in doctors' requirements for explanations, rooted in their training, experiences, and possibly even cultural contexts, exemplify the challenge. 

\subsection{Influence of explanation on doctor's decisions and multiplicity of valid diagnostic hypotheses}

Medical diagnoses are seldom straightforward. Multiple valid diagnostic hypotheses can exist for any given set of symptoms, with some being more probable than others. The data from our experiments demonstrated this multiplicity, with doctors agreeing with the different diagnostic hypotheses at experiment stages 1 and 3 for the same sets of patient symptoms. On the other hand, there is inherent noise in doctors' evaluations, as shown in cases where the same set of complaints and diagnoses produced varied evaluations when presented with different explanations.

\subsection{Performance of LLM in Generating Explanations}

From our findings, it is clear that LLMs:
\begin{itemize}
\item Are capable of producing high-quality explanations in most cases, further underscoring their potential in CDSS.
\item Generate explanations that can statistically significantly elevate doctor agreement rates with given diagnostic hypotheses, emphasizing their persuasive power.
\item Can err in a range of 5\% to 30\%, often due to the model's eagerness to align all symptoms with the current diagnostic hypothesis.
\end{itemize}

\subsection{Future Directions}

As we progress into an era where LLMs play a more pivotal role in healthcare, future work should focus on refining the interaction between LLMs and CDSS. Improving prompts and employing more advanced models can enhance the overall explanation quality. Furthermore, implementing additional checks to ensure the validity of generated answers will be crucial to ensure patient safety.

\subsection{Conclusion and Potential Impact}

The potential of Large Language Models in enhancing Clinical Decision Support Systems is undeniable. Their ability to produce human-like explanations can bridge the understanding gap between intricate medical diagnoses and healthcare professionals or patients. The present research is an example of this potential, outlining the strengths and challenges posed by integrating LLMs into healthcare. As we endeavor to enhance patient-centric care with technology, ensuring these models' judicious and ethical use becomes paramount. Hopefully, the insights from this research will serve as a stepping stone towards a future where technology and healthcare walk hand in hand, driving better patient outcomes.

\section{Competing interests}
No competing interest is declared.

\section{Author contribution statement}
DU devised the experiment, analyzed the results quantitatively, and wrote the manuscript; GZ organized and supervised the experiment, and AN did the qualitative analysis of the model errors.

\section{Acknowledgments}
This research did not receive external funding.

\bibliographystyle{unsrt}  
\bibliography{references}

\end{document}